# Generative Linguistics, Large Language Models, and the Social Nature of Scientific Success

Sophie Hao[a]

[a] Center for Data Science, New York University, USA <sophie.hao@nyu.edu>

Chesi's (*forthcoming*) target paper depicts a generative linguistics in crisis, foreboded by Piantadosi's (2023) declaration that "modern language models refute Chomsky's approach to language." In order to survive, Chesi warns, generativists must hold themselves to higher standards of formal and empirical rigor. This response argues that the crisis described by Chesi and Piantadosi actually has little to do with rigor, but is rather a reflection of generativists' limited social ambitions. Chesi ties the fate of generative linguistics to its intellectual merits, but the current success of language model research is social in nature as much as it is intellectual. In order to thrive, then, generativists must do more than heed Chesi's call for rigor; they must also expand their ambitions by giving outsiders a stake in their future success.

*1. Introduction*

To be a computational linguist is to inhabit two worlds—one seeking to explain why language is the way it is, the other trying to teach computers to read and write. Occasionally, these worlds collide. Chomsky (1968: 3) greeted the rise of computing technology with skepticism, arguing that "the kinds of structures that are realizable in terms of [computational methods] are simply not those that must be postulated to underlie the use of language." 55 years later, Piantadosi (2023: 15) celebrated the release of ChatGPT by directing that same criticism toward generative linguistics: "the success of large language models is a failure for generative theories because it goes against virtually all of the principles these theories have espoused."

Chesi (*forthcoming*) may not agree with Piantadosi's criticisms, but he does take them as a harbinger of scientific crisis. The minimalist program, hampered by a lack of formal and empirical rigor, has failed to produce a comprehensive, self-consistent theory of syntax. ChatGPT's apparent linguistic competence, in tandem with the success of computational accounts of gradient acceptability and online phenomena, seem to suggest that "generative linguistics no longer dictates the agenda for future linguistic challenges" (Chesi *forthcoming*: 2). In order to survive, Chesi warns, generativists need to make progress towards a theory that is based on precisely stated principles and evaluated on a common set of explananda.

Chesi's target paper presents the current collision of the worlds as a debate about the intellectual merits of generativist theories. According to Chesi, the success of generativism depends on generativists' ability to resolve their deficits of rigor, so that they can parry the theoretical attacks that language models have levied against core principles of minimalism. This response argues, contrary to Chesi's framing but consistent with current consensus in the history and sociology of science (Fleck 1935; Kuhn 1962; Mullins 1975; Latour 1984; Law & Lodge 1984), that the generativist crisis described by Piantadosi and Chesi is social in nature, and cannot be averted by intellectual means. To show this, I first review recent developments in language modeling research (§2), and then examine two debates that have pitted generative linguists against language model researchers (§3): the GRAMMAR *vs* PROBABILITY debate and the NATURE *vs* NURTURE debate. I argue that neither debate can end in intellectual victory because the disagreements between the two sides are grounded not in theoretical or empirical substance, but rather in the INCOMMENSURABILITY between their perspectives (Kuhn 1962; Feyerabend 1962). I then attribute Piantadosi's and Chesi's perceived decline of generativism to the narrowness of its social ambitions (§4): generativists mostly focus on explaining linguistic phenomena to themselves, whereas language model research has achieved social success by making itself relevant to a wide range of stakeholders. In order for generative linguistics to thrive, then, generativists must not only live up to their own ideals, but also convince society at large to invest in their future.

*2. A brief introduction to language models*

A LANGUAGE MODEL (**LM**) is a function that takes a sequence of TOKENS as input and predicts the next token in the sequence. The sequence is typically a text corpus, and the tokens are typically words, sub-word units, or orthographic characters. An LM's predictions are determined by PARAMETERS, whose values are set by TRAINING the LM on a corpus.

LMs have three primary applications in AI. Firstly, LMs are used to generate novel text, by repeatedly making next-token predictions. Secondly, LMs are used to estimate the probability that one sequence of tokens follows another. These sequence probabilities are used for tasks such as spelling correction and speech recognition

(Jurafsky & Martin 2024, Appendix B), as well as in psycholinguistic modeling (Hale 2001; Smith & Levy 2013). Thirdly, and most recently, 'large' LMs have been used as general-purpose AI models that perform arbitrary, user-defined tasks (see §2.2).

*2.1. Language models, past and present*

The invention of the LM is typically attributed to Shannon (1948), who describes what are now known as *n*-gram models. These models predict tokens based on CORPUS FREQUENCIES for sequences of length *n*. Current LM techniques do not use corpus frequencies, but instead rely on DEEP LEARNING (Cauchy 1847; Rumelhart *et al.* 1986; Kingma & Ba 2015), a framework that allows researchers to define and train arbitrarily complex probability models. There are two main kinds of deep learning LMs. RECURRENT NEURAL NETWORK models (**RNN**, Elman 1990; Hochreiter & Schmidhuber 1997; Cho *et al.* 2014) process sequences one token at a time, outputting predictions along with a description of the tokens they have seen so far. TRANSFORMER models (Vaswani *et al.* 2017) read sequences in parallel, with no *a priori* concept of linear order.

All three kinds of LMs are widely used, and each technique comes with advantages and limitations. *n*-gram models are cheap to train and execute, but they can only describe local co-occurrence relations between tokens. RNNs capture much more sophisticated linguistic relations than *n*-gram models (Elman 1991; Linzen *et al.* 2016; Gulordava *et al.* 2018; Marvin & Linzen, 2018; Conneau *et al.* 2018; Lakretz *et al.* 2019), but they have a mathematical bias towards shorter-distance dependencies (Pascanu *et al.* 2013). By eschewing linear order altogether, transformers overcome this bias while leveraging parallel computing techniques for efficient training. However, transformers underperform RNNs when less training data are available, and they require more time and memory for text generation than RNNs.

*2.2. 'Large' language models as general-purpose AI models*

Recent LM research has explored the idea of using LMs as general-purpose AI models, with potentially unlimited applications. The intuition is as follows. Consider the token sequence *The Italian translation of 'The cats saw the dogs.' is:*. Any high-performing LM will, *ipso facto*, predict the most likely continuation of this sequence to be some valid Italian translation of *The cats saw the dogs*. By this logic, a high-performing LM should be able perform any AI task, as long as the user can come up with a PROMPT that forces the LM to perform the task in order to predict the next token.

Radford *et al.* (2019b) and Brown *et al.* (2020) provide proofs of concept for this idea by attempting to train the highest quality LMs possible and prompt them to perform arbitrary tasks. The main technical challenge addressed by these studies is that of obtaining an LM that performs well enough to support prompting. To that end, Brown *et al.* (2020) create an LM that is as LARGE as possible, in terms of number of parameters, training corpus size, and computing resources used during training. These variables were identified by Kaplan *et al.* (2020) as being the most important determinants of LM performance, and studies have shown that largeness, in this sense, endows transformer LMs with general AI abilities (Brown *et al.* 2020; Zhang *et al.* 2021; Srivastava *et al.* 2023).

The CHATBOT ASSISTANTS that have entered consumer use in recent years, such as ChatGPT and Copilot, are the result of engineering improvements that have adapted Brown *et al.*'s (2020) LARGE LMs (**LLM**) to a commercial setting. These improvements include designing prompts to elicit complex forms of reasoning from LLMs (Nye *et al.* 2022; Wei *et al.* 2022b; Kojima *et al.* 2022), training LLMs to follow instructions in dialog format (Yi *et al.* 2019; Sanh *et al.* 2022; Wei *et al.* 2022a; Mishra *et al.* 2022), and making LLMs infer and comply with user intent while avoiding behavior that is unhelpful, socially harmful, or otherwise undesirable (Askell *et al.* 2021; Ouyang *et al.* 2022; Bai *et al.* 2022; Rafailov *et al.* 2023).

Despite this technological progress, LLMs face practical limitations that are intrinsic to the strategy of largeness. The vastness of LLM training corpora, open-endedness of LLM use cases, and technical complexity of certain LLM applications present challenges in evaluating the quality of LLMs and anticipating instances of undesirable behavior (Bender *et al.* 2021; Perez *et al.* 2022; Bowman *et al.* 2022). Furthermore, the logic of prompting does not guarantee that high-performing LMs will also perform well on general AI tasks. Not all information in training corpora is factually accurate (Trinh & Le, 2019; Ji *et al.* 2023; Huang *et al.* 2023), and larger LMs are more likely to quote misinformation from their training corpora (Lin *et al.* 2022; McKenzie *et al.* 2023).

*3. The futility of intellectual conflict*

The collision of worlds between generativists and LM researchers is often conceptualized as a debate regarding foundational assumptions of generative linguistics. Early generativist writings pointed to limitations of LMs as justification for a rationalist approach to the study of language (Chomsky 1956; 1957: Chapter 3). These arguments have been problematized by studies where LMs overcome said limitations (Pereira 2000; Mikolov 2012; and others). If generative linguistics is premised upon the limitations of

LMs, then it seems to follow that the success of LMs constitutes a foundational challenge to generativism.

This section argues that any attempt to win the foundational debate on generativism is futile, regardless of the evidence furnished by either side. This is because the generativist and LM-based approaches to language are INCOMMENSURABLE with one another (Kuhn 1962; Feyerabend 1962)—the two worlds operate from radically different conceptual frameworks, with distinct methods for making arguments and interpreting evidence. Incommensurability renders arguments from one world meaningless to the other, while also making it impossible to determine a winner impartially. In Milway's (2023: 1) words, "it is perhaps an axiom of criticism that one should treat the object of criticism on its own terms," but the foundational debate on generativism cannot be adjudicated while treating both sides on their own terms.

This section examines two instances of the foundational debate on generativism: the GRAMMAR *vs* PROBABILITY debate (§3.1) and the NATURE *vs* NURTURE debate (§3.2). I show that in both debates, generativists and LM researchers largely agree on substance, but differ in the rhetoric and methodologies used to instantiate that substance. The apparent tension between the two worlds therefore has little to do with facts or logic, but instead results from the mutual incompatibility of their perspectives.

*3.1. Grammar* vs *probability*

In arguing that grammar cannot be reduced to probability, Chomsky (1957: 16) famously asserts that "in any statistical model for grammaticalness, [*Colorless green ideas sleep furiously* and *Furiously sleep ideas green colorless*] will be ruled out on identical grounds as EQUALLY 'REMOTE' [emphasis added] from English" because "neither … has ever occurred in an English discourse." This claim of equal remoteness is clearly false, however, because most LMs assign higher probabilities to *Colorless green ideas* than *Furiously sleep ideas* (Pereira 2000; Mikolov 2012; Norvig 2017).

Equal remoteness presents an easy target for LMs to counterexemplify because, as Abney (1996) and Pereira (2000) point out, the notion of 'probability' presupposed above differs substantially from the assumptions that underlie LMs. If all unattested sentences have equal probability just because they have been observed the same number of times (zero), then it follows that probabilities are estimated from corpus frequencies, using entire sentences as tokens. These assumptions have never been used in LMs precisely because, as Chomsky argues, they are obviously unsuitable for language learning.

But even though LM-based counterexamples falsify the quotations above, such arguments miss the broader point of Chomsky's early writings. As explained in Chomsky

(1969: 53), Chomsky's caricatured model of probability is meant to represent "a narrowly Humean theory of language acquisition," a description that paper applies to Quine (1960). The idea is that a theory of language acquisition cannot account for language's "infinite use of finite means" (von Humboldt 1836) without a theory of how language acquirers solve the problem of induction (Plato 380 BCE; Hume 1739, 1748; Russell 1947). Chomsky's (1969: 57) criticism of Quine (1960) is that his theory does not pay sufficient attention to induction, and therefore predicts that a language "can … contain only the sentences to which a person has been exposed." Properly understood, then, the equal remoteness claim does not say that a probability model could never describe grammatical contrasts, but rather that it could not do so without an underlying theory of induction.

It thus becomes clear that Chomsky and his respondents are actually in agreement: as Baroni (2022) points out, all LMs do in fact implement some theory of induction. What appears to be a disagreement is the result of Chomsky's imprecise use of terms like 'probability' and 'statistical model.'

*3.2. Nature* vs *nurture*

According to Chomsky (1968: 24), the problem of induction in language ought to be explained by a theory of UNIVERSAL GRAMMAR (**UG**) that "tries to formulate the necessary and sufficient conditions that a system must meet to qualify as a potential human language." The POVERTY OF THE STIMULUS ARGUMENT (**POS**) attempts to derive facts about UG from observations about induction in language acquisition. One description of this logic, from Pullum & Scholz (2002), is reproduced below (though see Clark & Lappin 2011: Chapter 2; Berwick *et al.* 2011; Lasnik & Lidz 2016; and Pearl 2022 for other overviews).

(1) Poverty of the stimulus argument, Pullum & Scholz's (2002) version
   a. Human infants learn their first languages either by data-driven learning or by innately-primed learning.
   b. If human infants acquire their first languages via data-driven learning, then they can never learn anything for which they lack crucial evidence.
   c. But infants do in fact learn things for which they lack crucial evidence.
   d. Thus human infants do not learn their first languages by means of data-driven learning.
   e. Conclusion: human infants learn their first languages by means of innately-primed learning.

LMs are used in a style of counterargument to the POS that I call the EXISTENCE PROOF REBUTTAL (**EPR**, see Linzen & Baroni 2021: §6.1). The EPR uses an LM or some other 'data-driven' machine learning model as a counterexample to (1b). For example, Wilcox *et al.*'s (2024) EPR argument tests LMs' knowledge of island constraints (Ross 1967), with the idea that UG hypotheses meant to derive island constraints can only be justified by the POS if they account for facts that LMs fail to learn.

Here I discuss three points of incommensurability between generativists and LM researchers that make the EPR logically incompatible with the POS.

Firstly, it is unclear what the difference is between 'data-driven learning' and 'innately-primed learning.' Although deep learning algorithms are typically assumed to be data-driven, these algorithms are not *tabulae rasae*. As Baroni (2022: 7) argues, the mathematical structure of deep learning models "[encodes] non-trivial structural priors facilitating language acquisition and processing" that limit the range of patterns they can capture (Hao *et al.* 2022; Merrill *et al.* 2022; Yang *et al.* 2024) and influence their preference for certain generalizations over others (McCoy *et al.* 2018; Kharitonov & Chaabouni 2021). The random initial values assigned to RNN and transformer parameters are also thought to encode random priors, such that larger models have a greater chance of accidentally encoding priors that facilitate language learning (Frankle & Carbin 2019; Chen *et al.* 2020; Prasanna *et al.* 2020). On the other hand, human learners are clearly driven by data even if we assume they are innately-primed, though there is still no explicit theory of how this works apart from a few notable proposals (Wexler & Culicover 1980; Pinker 1984). Premise (1a) of the POS, then, may very well be a false dichotomy.

Secondly, POS and EPR arguments typically use incommensurable notions of linguistic knowledge. The POS arguments surveyed by Pullum & Scholz (2002) follow Chomsky (1957; 1995) in accessing linguistic knowledge through judgments of syntactic and/or semantic well-formedness. But LMs have no concept of well-formedness, so their linguistic knowledge is instead defined by their ability to rank minimal pairs of well-formed and ill-formed sentences through their probability estimates (Elman 1990; Linzen *et al.* 2016; Warstadt *et al.* 2020; Gauthier *et al.* 2020). One implication of this discrepancy concerns the question of whether negative evidence is 'crucial' for language acquisition. As Crain & Nakayama (1987: 527) articulate, positive evidence could only cause rules to be "added to children's grammars; [but] such addition would not necessarily prompt the abandonment of [incorrect rules]." Abandonment of incorrect rules may be crucial for acquiring well-formedness judgments, but it is not necessary for acquiring minimal-pair contrasts: to rank minimal pairs, an LM never needs to conclude that unattested rules are incorrect, only that they are less likely to occur in corpora than attested rules. The relevance of Crain and Nakayama's argument to the nature *vs* nurture debate is therefore contingent on how the concept of linguistic knowledge is operationalized.

Thirdly, as Clark & Lappin (2011: §2.4) point out, POS arguments often focus on contrasts motivated by specific, 'theory-internal' hypotheses about learners, which may not apply to LMs. To illustrate, consider the following minimal pairs.

(2) Long-distance agreement (Linzen *et al.* 2016)
    a.    The keys to the cabinet are on the table.
    b.    * The keys to the cabinet is on the table.

(3) Auxiliary inversion (Chomsky 1968)
    a.    Will the subjects who will act as controls be paid?
    b.    * Will the subjects who act as controls will be paid?

In (2–3), the 'a' sentences resemble naturally occurring sentences, while the 'b' sentences are unnatural mistakes that learners might make under certain assumptions. Unless data-driven learners satisfy these assumptions, we should expect them to prefer the 'a' sentences by default, since they are more likely to occur in corpora. For example, (2b) is a plausible mistake for LMs that are biased against long-distance dependencies, like *n*-gram models and RNNs, while (3b) is meant to be a mistake characteristic of learners that induce movement operations without a 'structural' bias. Since most LMs have no concept of movement, we have no reason to expect LMs to prefer (3b) over (3a); an LM's knowledge of (3) is therefore evidence only of the null hypothesis, and not of, say, an EPR argument against a UG principle codifying structure-sensitivity.

In summary, neither side of the nature *vs* nurture debate has a precise theory of what distinguishes 'nature' from 'nurture'; indeed, all learners are both 'innately-primed' and 'data-driven' by logical necessity. Where the two sides differ is in their assumptions about how linguistic knowledge is accessed, and what kinds of mistakes data-driven learners are likely to make without innate priming. These differences have more to do with relatively arbitrary metatheoretical choices than with the tension between nature and nurture, but they have profound implications for what arguments are relevant, what phenomena require explanation, and how evidence ought to be interpreted. It is therefore these metatheoretical choices that are in contention when trying to referee the nature *vs* nurture debate, as Chesi (*forthcoming*) does in §3.1. But while it is plausible to think that the question of nature *vs* nurture ultimately has an answer, it is unclear whether any truth is to be found in a conflict between two incommensurable approaches to science.

*4. The social nature of scientific success*

Why did Piantadosi wait until 2023 to declare the end of generative linguistics, when OpenAI had publicized its text generation capabilities as early as 2019 (Radford *et al.* 2019a)? The obvious answer is that the impetus for Piantadosi's manuscript was not the intellectual advancements that led to ChatGPT, but rather the social success that OpenAI and other companies have enjoyed by commercializing this technology. The same was true in prior collisions of the worlds: Abney (1996), Pereira (2000), and Norvig (2017) all cite the practical, not just intellectual, success of data-driven methods as reasons to take them seriously.

This section argues that limited social ambition is at the root of Piantadosi's and Chesi's generativist crisis, and that greater social ambition can help prevent it. Indeed, the "end of (generative) linguistics as we know it" (Chesi *forthcoming*: 1) is not a world where generativist theories are incorrect, but rather a world where generative linguistics is no longer practiced. I start by arguing that generative linguistics has traditionally thought of itself as a strictly intellectual enterprise, whereas the current scholarly consensus is that science is an intrinsically social activity (§4.1). I then argue that a narrow focus on intellectual goals is actually counterproductive to those goals, while social openness can be key to fulfilling Chesi's (*forthcoming*) call for rigor (§4.2).

*4.1. The explanatory telos: a culture of strict intellectualism*

Generative linguistics is characterized by an EXPLANATORY TELOS: generativists take it for granted that the singular purpose of generative linguistics is to explain language to generativists. Kodner *et al.* (2023: 12–13) articulate this attitude explicitly, dismissing Piantadosi (2023) on the grounds that "the role of a scientific theory is to ELUCIDATE AND EXPLAIN [(Popper 1934)], and LLMs largely fail to do either." Kodner *et al.* go on to deny that research done by the private sector should even count as science, because "the goals of [corporations] … (i.e., to increase profits) are not the same as the goals of [scientists] … (i.e., to come to a scientific understanding of language)."

The explanatory telos is invoked in Chomsky's public commentary on LLMs (Marcus 2002; Chomsky *et al.* 2023), and presupposed by Chesi's (*forthcoming*) extensive discussion of explanatory adequacy. The origins of the explanatory telos, in texts like Chomsky (1957: Chapter 8; 1965: Chapter 1; 1966; 1968: Chapter 1), can be understood as a reaction to structuralist, behaviorist, and LM-based approaches to language, which, in Chomsky's view, had thitherto focused on description and modeling of superficial patterns while neglecting explanation.

That the explanatory telos is common sense for many linguists today is a testament to the success that generativism has enjoyed over the past 68 years. But it would be a mistake to attribute generativism's past or future success exclusively to its fulfillment of the explanatory telos, as Kodner *et al.*'s and Chesi's arguments suggest. Science does not exist outside of social and historical context (Fleck 1935; Kuhn 1962), and neither does scientific success (Mullins 1975; Law & Lodge 1984). For example, the existence of microbes has been key to explaining such biological phenomena as fermentation and contagious disease (Pasteur 1876: Chapter VI; Pasteur *et al.* 1878; Pasteur 1880). But as Latour (1984) argues, the success of microbe theory cannot be accounted for by its explanatory power alone, without reference to the practical utility it provided to doctors, public health officials, and colonial authorities in 19th century France. As for generative linguistics, much of its early development and proliferation is owed to the United States' abundant defense spending on university expansion and relatively unrestricted research grants during the 1960s (Newmeyer & Emonds 1971; McCawley 1976; Koerner 1983; Murray 1994: Chapter 9).

There is, of course, nothing wrong with the pursuit of explanation and understanding. The point of this subsection is that the explanatory telos is neither a complete characterization of the scientific enterprise, nor an adequate theory of scientific success. Generative crisis therefore cannot be averted without addressing the social nature of science.

*4.2. Symbiosis of the social and the intellectual*

Generativists need not be concerned with social success, however, in order to benefit from a more socially ambitious generativism. Here I give three reasons why social openness can enhance the intellectual merits of generativism, while a narrow focus on the explanatory telos can be counterproductive to that end.

First, a narrow focus on the explanatory telos limits generativists' funding opportunities. Much of the defense funding that supported early generativist work at MIT was granted not for furthering the explanatory telos, but for conducting research on machine translation (Koerner 1983; Knight 2016: Chapter 8). As Newmeyer (1986: 12–13) reports, when the United States government began "demanding demonstrable military relevance for all military spending, such funding came to a complete halt." In one documented instance, the Ford Foundation "[refused] to fund generative research" because "it objects to the fact that generativists 'have isolated [themselves] from the world of non-linguistic events and concentrated on abstract and formal theories about the nature and structure of language' (Fox & Skolnick 1975)."

Second, I claim that the explanatory telos is structurally responsible for Chesi's (*forthcoming*) FORMALIZATION and EVALUATION ISSUES. It has been argued that scientific explanations are subject to pragmatics (van Fraassen 1980: Chapter 6; Achinstein 1985; Hao 2022: Chapter 2), and that subjective evaluation of explanatory power depends on an explanation's ability to instill understanding (Waskan *et al.* 2014; Wilkenfeld 2014; Wilkenfeld & Lombrozo 2020). By simplifying linguistic analyses, formal vagueness and data idealization can enhance a reader's feeling of understanding, thereby increasing a theory's perceived explanatory power. The explanatory telos thus introduces pragmatic incentives to seek rhetorical clarity at the expense of rigor, effectively balancing Grice's (1975) maxim of manner against the maxim of quality. Indeed, according to Chesi (*forthcoming*: 30), generativists' dismissal of computational methods consists in large part of "skepticism towards the inherent complexity of [such] methods."

Third, non-academic institutions conduct important research that academic institutions cannot, because of resource limitations, or will not, because such research does not fulfill the explanatory telos. LLMs, for instance, are generally considered too expensive for academic institutions to train from scratch (Strubell *et al.* 2019; Sathish *et al.* 2024; Cottier *et al.* 2025), but their impact on academic research has been enormous. In addition to instigating the study of prompting and largeness, LLMs have also helped refine the POS (Baroni 2022; Katzir 2023; Wilcox *et al.* 2024; Chesi *forthcoming*) and further our understanding of psycholinguistic processing during reading (Wilcox *et al.* 2020; Hao *et al.* 2020; Oh & Schuler 2023), among other intellectual contributions. Specifically regarding Chesi's (*forthcoming*) call for rigor, it should be noted that the most thoroughly fleshed-out syntactic formalisms are TREEBANKS consisting of structural analyses for thousands of sentences in many languages, rendered in frameworks like HPSG (Pollard & Sag 1994; Oepen *et al.* 2002), CCG (Steedman & Baldridge 2006; Hockenmaier & Steedman 2007; Tran & Miyao 2022), and even MG (Stabler 1997; Torr 2017, 2018). These treebanks are developed by computational linguists, who use them in traditional AI approaches to language. Similarly, in economics, the most detailed and comprehensive models of the global macroeconomy are maintained by the United States Federal Reserve Board in order to inform fiscal and monetary policy (Brayton *et al.* 2014). As Chesi (*forthcoming*) argues, large-scale projects like these are prerequisites for rigor, but difficult to justify from the explanatory telos. Collaborators with different priorities can therefore play a crucial role in bringing such projects to fruition.

*5. Conclusion*

The thesis of this response is that Chesi's (*forthcoming*) account of generativist crisis and survival is based an outdated understanding of science as a purely intellectual enterprise, where success is determined by purely intellectual factors. Historians and sociologists have shown that science is in fact a social enterprise, and that social factors are crucial determinants of scientific success. That the IUSS Pavia roundtable did not produce a syntactic theory of everything, or that the foundations of minimalism are still under debate, is just as much reason to continue working on generative linguistics as reason to abandon it. Which of these possible worlds becomes reality depends on whether future linguists find the generativist enterprise worthwhile. Intellectual progress along these dimensions can certainly help, but such achievements alone will mean little to young scholars driven by a sense of social purpose, or theoreticians struggling to secure a tenure-track position.

The rise of computational methods is not unique to linguistics, but is rather a broad social and intellectual trend that has been called "the end of theory" (Anderson 2008; Hey *et al.* 2009; Kitchin 2014). AI scholars recently had their own reckoning with the end of theory; the unreasonable effectiveness of "general methods that leverage computation" has been described as a "bitter lesson" by those sympathetic to the explanatory telos (Sutton 2019). Nevertheless, theory continues to thrive when it is uniquely positioned to offer social value. Theory drives the models that inform monetary (e.g., Brayton *et al.* 2014) and environmental policy (e.g., Voldoire *et al.* 2013; Zhao *et al.* 2018a,b). Theory is used to forecast the macroeconomic impact of AI (e.g., Acemoglu 2024) and analyze ongoing military conflicts (e.g., Edinger 2022; İdrisoğlu & Spaniel, 2024). Theoretical analysis of AI model outputs is even legally required in some instances, under 'right to explanation' laws in the United States and the European Union. Unlike generativist theories, which are written for the benefit of generativists, these theories are written for external audiences, who then become stakeholders in their success.

The explanatory telos notwithstanding, linguists have made tremendous contributions to our understanding of socially pertinent issues like discrimination (e.g., Purnell *et al.* 1999; Rickford & King 2016), pop culture (e.g., Kawahara *et al.* 2018), and social media (e.g., Crystal 2011; McCullough 2019). Generativist ideas in particular take center stage in Bjorkman's (2017) analysis of English personal pronouns of non-binary gender. The world is waiting for generative linguistics to succeed. All we have to do is accept its invitation.

*Acknowledgements*

I thank Dana Angluin, Cristiano Chesi, Robert Frank, Michael Y. Hu, Niels Lee, Cara Su-Yi Leong, Tal Linzen, William Merrill, Byung-Doh Oh, Timothy O'Donnell, Jackson Petty, Samuel Schotland, William Timkey, and Lindia Tjuatja for discussion and feedback, and Pier Marco Bertinetto for his invitation to write this paper.